\documentclass[runningheads]{llncs}
\usepackage[T1]{fontenc}
\usepackage{graphicx,verbatim,xcolor,array,caption,booktabs,tabularx,ragged2e,listings,amsmath,amssymb,multirow}
\usepackage{hyperref}
\usepackage[capitalize]{cleveref}
\begin{document}
\title{Med-Art: Diffusion Transformer for 2D Medical Text-to-Image Generation}

\author{Changlu Guo\and
Anders Nymark
Christensen\and
Morten Rieger
Hannemose}
% \author{Changlu Guo\orcidID{0000-0002-8699-0564} \and
% Anders Nymark
% Christensen\orcidID{0000-0002-3668-3128} \and
% Morten Rieger
% Hannemose\orcidID{0000-0002-9956-9226}}
%
\authorrunning{Guo et al.}

\institute{Department of Applied Mathematics and Computer Science, \\
Technical University of Denmark, Kgs. Lyngby, Denmark \\
\email{\{chagu, anym, mohan\}@dtu.dk}}

% \author{Anonymized}  %% Added for anonymized MICCAI 2025 submission
% \authorrunning{Anonymized Author et al.}
% \institute{Anonymized Affiliations \\
%     \email{email@anonymized.com}}

\maketitle              % typeset the header of the contribution
\begin{abstract}
Text-to-image generative models have achieved remarkable breakthroughs in recent years. However, their application in medical image generation still faces significant challenges, including small dataset sizes, and scarcity of medical textual data. To address these challenges, we propose Med-Art, a framework specifically designed for medical image generation with limited data. Med-Art leverages vision-language models to generate visual descriptions of medical images which overcomes the scarcity of applicable medical textual data. Med-Art adapts a large-scale pre-trained text-to-image model, PixArt-$\alpha$, based on the Diffusion Transformer (DiT), achieving high performance under limited data. Furthermore, we propose an innovative Hybrid-Level Diffusion Fine-tuning (HLDF) method, which enables pixel-level losses, effectively addressing issues such as overly saturated colors. We achieve state-of-the-art performance on two medical image datasets, measured by FID, KID, and downstream classification performance. The project is available at \url{https://medart-ai.github.io}.

\keywords{Text-to-image \and generative models  \and medical image generation.}
% Authors must provide keywords and are not allowed to remove this Keyword section.

\end{abstract}
%
%
% DMs outperform traditional generative adversarial networks (GANs) in terms of image quality, diversity, and training stability~\cite{dhariwal2021diffusion,rombach2022high}.
\section{Introduction}
The emergence of diffusion models~\cite{ho2020denoising} has driven huge advancements in text-to-image generation, achieving unprecedented success in natural image synthesis. Proprietary models such as Imagen 3~\cite{baldridge2024imagen}, alongside open-source frameworks like Stable Diffusion~\cite{podell2023sdxl}, Hunyuan-DiT~\cite{li2024hunyuan}, and PixArt-$\alpha$~\cite{chen2023pixart}, have demonstrated remarkable capabilities in generating high-fidelity and diverse images. These models are trained on web-scale data, which contains little medical data, greatly limiting their ability to generate this out-of-the-box. Despite this, diffusion-based generative models hold significant potential in medical contexts, such as synthesizing rare diseases, enabling privacy-preserving data sharing, and improving medical education.
\begin{figure}[t]
  \centering
   \includegraphics[width=1\linewidth]{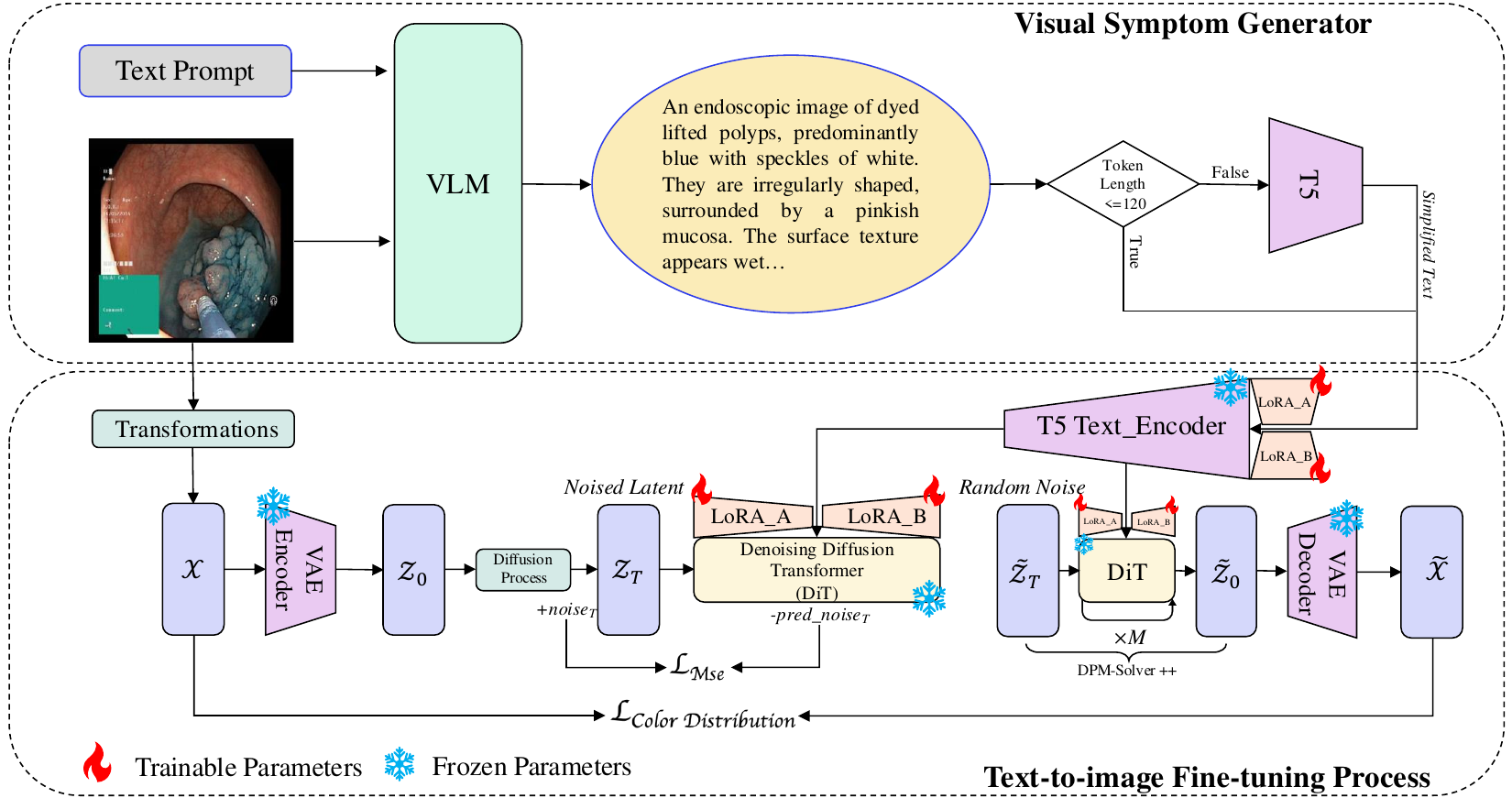}

   \caption{The architecture of Med-Art (the parameters of DiT are shared)
   }
   \label{fig:medart}
\end{figure}

Recent studies have explored applications of diffusion models across diverse medical scenarios, including 3D vascular graph synthesis~\cite{prabhakar20243d}, iterative online image synthesis for mitigating class imbalance~\cite{li2024iterative}, and anatomically constrained generation using segmentation masks~\cite{konz2024anatomically}. However, unlike the rapid advancements in natural image generation, research on medical domain-specific text-to-image diffusion models remains less explored. Existing methods, such as RoentGen~\cite{bluethgen2024vision}, fine-tune Stable Diffusion to generate chest X-ray images but are constrained by their reliance on radiology-specific text prompts. For most other medical imaging modalities, such detailed textual descriptions are unavailable. For instance, Akrout et al.~\cite{akrout2023diffusion} generate skin lesion images using only class labels as prompts. This constraint restricts the model’s ability to fully exploit the potential of text-to-image technology for medical image generation. 

Therefore, the application of text-to-image within the medical domain still faces significant bottlenecks. This is largely due to strict privacy and ethical constraints that severely limit public access to medical images and their associated textual data. Furthermore, text connected to medical images is typically highly specialized, such as medical records, which require specialized knowledge to connect to images. Additionally, we observed that fine-tuning models pretrained on natural images can cause color oversaturation issues in color-sensitive 2D RGB modalities such as endoscopic and dermoscopic images.
To overcome these issues, we propose Med-Art, a framework designed to adapt and enhance text-to-image models for 2D medical image generation, improving fidelity, accuracy, and clinical applicability. Our main contributions are:
% \begin{itemize}
%     \item \textbf{Visual Symptom Generator (VSG)}: Leveraging Vision-Language Models (VLMs)~\cite{vlm} to generate high-quality, contextually relevant medical descriptions, mitigating textual data scarcity and aiding downstream T2I fine-tuning.

%     \item \textbf{Efficient Fine-Tuning of PixArt-$\alpha$}: Applying Parameter-Efficient Fine-Tuning (PEFT) techniques like LoRA~\cite{lora} to both the Diffusion Transformer (DiT)~\cite{dit} and Text Encoder, enabling effective adaptation to medical semantics with minimal computational cost.

%     \item \textbf{Hybrid-Level Diffusion Fine-Tuning (HLDF)}: Addressing severe color oversaturation by integrating advanced numerical solvers~\cite{dpmsolver}\cite{dpmsolver++} into fine-tuning, allowing pixel-level loss computation and enhancing generation quality.

%     \item \textbf{Superior Performance}: Med-Art achieves SOTA results on two small-scale medical datasets, significantly outperforming Stable Diffusion and PixArt-$\alpha$ in FID/KID scores and downstream classification tasks.
% \end{itemize}
% To explore this issue, we propose the Med-Art framework, focusing on adapting and enhancing existing SOTA T2I generative model to meet the specific needs of the medical field while improving the fidelity, accuracy, and applicability of generated medical images. Our main contributions are as follows:
\begin{itemize}
    \item This is the first work to leverage a vision-language model (VLM)~\cite{zhang2024vision} to generate high-quality, visually grounded natural image descriptions that resemble the language distribution seen during text-to-image model pretraining, thereby addressing the scarcity of medical textual data and enabling effective text-to-image fine-tuning.

    % \item Based on the Diffusion Transformer (DiT)~\cite{peebles2023scalable} model PixArt-$\alpha$, we employ LoRA~\cite{hu2021lora}, to both the DiT module and the text encoder. This approach optimizes only a small subset of parameters, enabling the model to effectively capture specific medical semantics and generate clinically relevant, high-quality images under limited data.
    \item Based on the Diffusion Transformer (DiT) [23] model PixArt-$\alpha$, we apply LoRA [13] to both the DiT module and text encoder, optimizing only a small subset of parameters. This enables the model to capture medical semantics and generate high-quality medical images from limited data.

    % \item Based on the Diffusion Transformer (DiT)~\cite{peebles2023scalable} model PixArt-$\alpha$, we employ LoRA~\cite{hu2021lora} for parameter-efficient fine-tuning of both the DiT and text encoder, facilitating effective adaptation to 2D medical text-to-image generation tasks. 
    % Our ablation experiments show that LoRA achieves better performance than the latest DoRA~\cite{liu2024dora} method in this task.
    \item 
    %We observe severe color distortion issues, such as oversaturation, in generated medical images, which affect clinical utility and semantic consistency. 
    %To address this problem, we propose Hybrid-Level Diffusion Fine-Tuning (HLDF), 
    We propose a novel method to ensure color consistency by integrating the Diffusion Probabilistic Model Solver~\cite{lu2022dpm++} into generative model fine-tuning. Generating images during training enables pixel-level loss computation.
\end{itemize}

 Experimentally, Med-Art achieves state-of-the-art performance on two small-scale medical image datasets, significantly surpassing existing methods such as the Stable Diffusion series and PixArt-$\alpha$, and also demonstrates superior downstream classification task performance.

\section{Method}
Med-Art is structured, as shown in \cref{fig:medart}. The Visual Symptom Generator (VSG) generates detailed visual text descriptions from the original medical images and prompts. These descriptions, combined with the transformed images (e.g., normalized), are used in the fine-tuning process to learn the relationship between visual representations and text.

\begin{table}[hbt]
    \caption{A single sample from both datasets, along with a simple caption, and a caption from the VSG, used for the text-to-image model.}\label{table1}
    \centering
    \begin{tabularx}{\textwidth}{p{1.70cm}>{\centering\arraybackslash}X >{\centering\arraybackslash}X} 
    
        \toprule
        Dataset& Kvasir & Skin lesions \\ 
        \midrule
        Sample & \raisebox{-0.5\height}{\includegraphics[width=2.2cm]{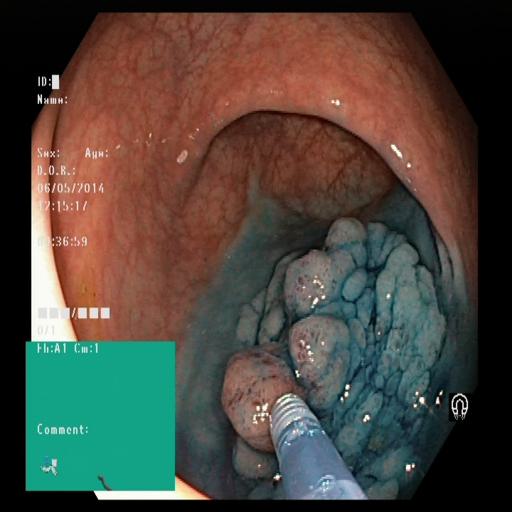}} & \raisebox{-0.5\height}{\includegraphics[width=2.2cm]{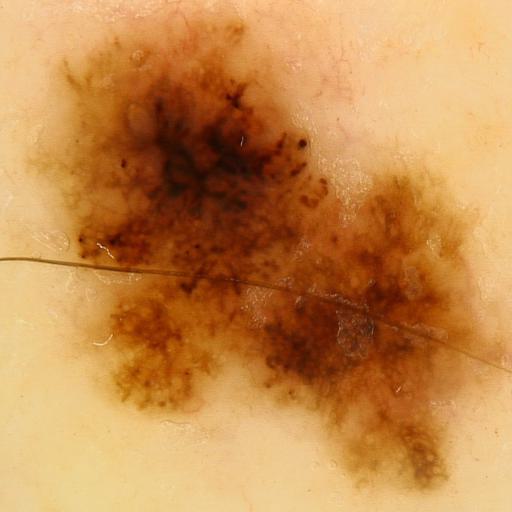}} \\ 
        \midrule
        Modality& endoscopic & dermoscopic \\ 
                \midrule
        Category& dyed lifted polyps & Melanoma \\ 
        \midrule
\parbox[c][0.8cm][c]{1.7cm}{Visual \mbox{features}}  & 
\parbox[c][0.8cm][c]{5cm}{\fontsize{8}{8}\selectfont specific color, shape, surrounding mucosa, surface texture, bleeding condition, and other relevant details} & 
\parbox[c][0.8cm][c]{5cm}{\fontsize{8}{8}\selectfont specific color, symmetry, border, shape, texture and dermoscopic patterns} \\

        \midrule
\fontsize{8}{8}\selectfont
\parbox[c][0.8cm][c]{1.7cm}{Simple caption}   & 
\parbox[c][0.5cm][c]{\linewidth}{\centering  \fontsize{8}{8}\selectfont An endoscopic image of dyed lifted polyps} & 
\parbox[c][0.5cm][c]{\linewidth}{\centering \fontsize{8}{8}\selectfont A dermoscopic image of Melanoma} \\ 

        \midrule
        \parbox[c][2.5cm][c]{1.7cm}{LLaVA-Next refined caption (VSG)} &
        \parbox[c][2.5cm][c]{\linewidth}{\fontsize{9}{10}\selectfont An endoscopic image of dyed lifted polyps, predominantly blue with speckles of white. They are irregularly shaped, surrounded by a pinkish mucosa. The surface texture appears wet and smooth. There is no active bleeding evident in the image.} 
        & \parbox[c][2.5cm][c]{\linewidth}{\fontsize{8}{8}\selectfont A dermoscopic image of Melanoma shows a brownish-black irregularly shaped patch with a rough, granular texture and a wavy border. It has a central dark area with lighter satellite lesions and a radial streak pattern emanating from the central lesion. The background is a pale, translucent skin tone.} \\ 
        \bottomrule
    \end{tabularx}
\end{table}

\subsection{ Visual Symptom Generator (VSG)}
In the medical domain, due to the specialized nature and privacy concerns, obtaining image descriptions with text annotations is challenging. Our VSG leverages the knowledge and multimodal capabilities of vision-language models to generate visual descriptions of medical images. Specifically, as shown in the upper part of \cref{fig:medart}, we use the LLaVA-Next~\cite{li2024llava} model to generate descriptions for medical images through prompts. This allows for the description of the visual appearance in a more natural language, which is better suited for the pretrained text-to-image models. To generate the corresponding visual symptom descriptions, we use the image and the following structured textual prompt:
\begin{quote}
Q: An/A \texttt{\{modality\}} image of \texttt{\{category\}}. Please describe the \texttt{\{modality\}} image of \texttt{\{category\}} using the following visual features: \texttt{\{visual features\}}. Start with `An/A \texttt{\{modality\}} image of \texttt{\{category\}}' and ensure the description does not exceed 100 words.
\end{quote}
Here \texttt{\{modality\}} represents the imaging modality of the dataset, such as dermoscopic for skin images; \texttt{\{category\}} refers to a specific category, and \texttt{\{visual features\}} denotes the corresponding visual features associated with the given modality. This information and examples of generated texts are in \cref{table1}. The pre-trained DiT from PixArt-${\alpha}$ is limited to 120-tokens, yet certain generated texts exceed this constraint, leading to information loss upon truncation; thus, we employ T5~\cite{raffel2020exploring} for text simplification, directly retaining texts within 120 tokens while prompting T5 with \texttt{"simplify: \{generated\_text\}"} for longer texts to generate a concise 90-120 token version, ensuring information integrity.
\subsection{Text-to-image Fine-tuning Process}
\textbf{Training Procedure for PixArt-$\alpha$} PixArt-$\alpha$ builds upon conditional latent diffusion~\cite{rombach2022high}, using a pretrained encoder~\cite{esser2021taming} to encode the input image \(x\) into a latent representation \(z_0 = E(x)\). The model learns the conditional probability distribution \(p(z_0|y)\), where \(y\) is the text describing the target image. During training, at each time step \(t\)  (uniformly sampled from \(\{1, \dots, T\}\)), a noisy latent variable is generated as:
\(
z_t = \sqrt{\bar{\alpha}_t} z_0 + \sqrt{1 - \bar{\alpha}_t} \epsilon,
\)
where \(\bar{\alpha}_t\) is the cumulative noise attenuation coefficient, and \(\epsilon \sim \mathcal{N}(0, I)\) represents Gaussian noise. The conditional denoising network \(\epsilon_\theta(z_t, t, \tau_\theta(y))\) is trained to predict \(\epsilon\), optimizing the objective:
\begin{equation}
\mathcal{L}_{\text{PixArt-$\alpha$}} := \mathbb{E}_{E(x),y,\epsilon \sim \mathcal{N}(0,I),t} \left[\|\epsilon - \epsilon_\theta(z_t, t, \tau_\theta(y))\|_2^2\right]
\end{equation}
Here, \(\tau_\theta(y)\) is the text embedding produced by a frozen text encoder (e.g., T5 \cite{raffel2020exploring}), which guides denoising and ensures consistency. The denoising network, built on DiT, integrates \(\tau_\theta(y)\) with latent features \(\phi(z_t)\) using cross-attention~\cite{vaswani2017attention}, enabling conditional generation.

\noindent \textbf{LoRA}
% To achieve efficient domain adaptation of T2I generation models for the medical domain, we actively integrate LoRA into PixArt-$\alpha$. LoRA fine-tunes the Text Encoder and denoising network using low-rank decomposition, allowing domain-specific optimization while keeping the original pretrained weights frozen. Unlike traditional methods that freeze the Text Encoder, LoRA adjusts the Text Encoder to better handle medical terminology and contexts, enhancing text embeddings for multimodal fusion. The updated text embeddings are expressed as:
To efficiently adapt text-to-image generation models to the medical domain, we integrate LoRA~\cite{hu2021lora} into PixArt-$\alpha$. 
% LoRA fine-tunes the text encoder and denoising network using low-rank decomposition, enabling domain-specific optimization while keeping pretrained weights frozen. 
Unlike previous methods that freeze the text encoder \cite{chen2023pixart,li2024hunyuan}, LoRA adjusts the text encoder to better handle medical context, improving text embeddings. The updated text embeddings are expressed as:
\(
\tau_{\theta + \Delta\theta}(y) = \tau_\theta(y) + \Delta\tau(y), \quad \Delta\tau(y) = W_A^{(T5)} W_B^{(T5)} y\),
where \( W_A^{(T5)} \in \mathbb{R}^{d \times r} \) and \( W_B^{(T5)} \in \mathbb{R}^{r \times d} \) are low-rank matrices, with \( r \ll d \), ensuring efficient updates with minimal additional parameters. In the denoising network DiT, LoRA is applied to key components, including attention mechanisms and feed-forward layers. For instance, the query matrix in the cross-attention mechanism is updated as: \( W_Q' = W_Q + W_A^{(DiT)} W_B^{(DiT)}\),
with similar updates for the key and value matrices. These adjustments enable better integration of medical text conditions into the latent space, enhancing both semantic consistency and detail accuracy. With these modifications, the denoising network in PixArt-$\alpha$ is redefined as:
\begin{equation}
\epsilon_{\theta + \Delta\theta}(z_t, t, \tau_{\theta + \Delta\theta}(y)) = \epsilon_\theta(z_t, t, \tau_{\theta + \Delta\theta}(y)) + \Delta\epsilon(z_t, t, \tau_{\theta + \Delta\theta}(y))
\end{equation}
where \(
\Delta\epsilon(z_t, t, \tau_{\theta + \Delta\theta}(y)) = W_A^{(DiT)} W_B^{(DiT)} \phi(z_t, t, \tau_{\theta + \Delta\theta}(y)) \). The optimization objective is then reformulated as:
\begin{equation}
\mathcal{L}_{\text{Pixart-$\alpha$+LoRA}} := \mathbb{E}_{E(x), y, \epsilon \sim \mathcal{N}(0,I), t} \left[\|\epsilon - \epsilon_{\theta + \Delta\theta}(z_t, t, \tau_{\theta + \Delta\theta}(y))\|_2^2\right]
\end{equation}

\noindent \textbf{Hybrid-Level Diffusion Fine-tuning (HLDF)}
The fine-tuning described in the previous sections operates purely in the latent space, but is prone to generating images with highly saturated colors. To address this, we penalize deviations of the color distribution in pixel space. 
We minimize the difference in the mean (\(\mu_c\)) and standard deviation (\(\sigma_c\)) of each color channel \( c \) between the input image \( x \) and the generated image \( \tilde{x} \). The loss is defined as:
\newcommand{\lcolordist}{\mathcal{L}_{\text{color}}}
\begin{equation}
\label{eq:colordist}
\lcolordist := \mathbb{E}_x\Bigl[\sum_{c} \Bigl(\|\mu_c(\tilde{x}) - \mu_c(x)\|_2^2 + \|\sigma_c(\tilde{x}) - \sigma_c(x)\|_2^2\Bigr)\Bigr]
\end{equation}
In order to compute this loss, we need to generate images during training. For this, we use classifier-free guidance~\cite{ho2022classifier} with a guidance weight of 4.5, and for sampling, we utilize the DPM-Solver++~\cite{lu2022dpm++}, which enables the generation of high-quality images in 15–20 steps.
\begin{comment}
, which is defined as:
\begin{equation}
\epsilon_{\text{CFG}}(z_t, t, \mathbf{e}_y) = \epsilon_{\theta + \Delta\theta}(z_t, t, \emptyset) + w \cdot \Bigl(\epsilon_{\theta + \Delta\theta}(z_t, t, \mathbf{e}_y) - \epsilon_{\theta + \Delta\theta}(z_t, t, \emptyset)\Bigr)
\end{equation}
where \( w = 4.5 \), controls the influence of conditional information, and \( \mathbf{e}_y = \tau_{\theta + \Delta\theta}(y) \) represents the condition embedding generated from the input text \( y \). 
By combining conditional guidance with noise predictions, CFG starts from the initial Gaussian noise \( \tilde{z}_{t_0} = \tilde{z}_T \sim \mathcal{N}(0, I) \) and iteratively optimizes over \( M \) steps to produce the target latent variable \( \tilde{z}_0 = \tilde{z}_{t_M} \).
\end{comment}

\begin{figure}[t]
  \centering
   \includegraphics[width=1\linewidth]{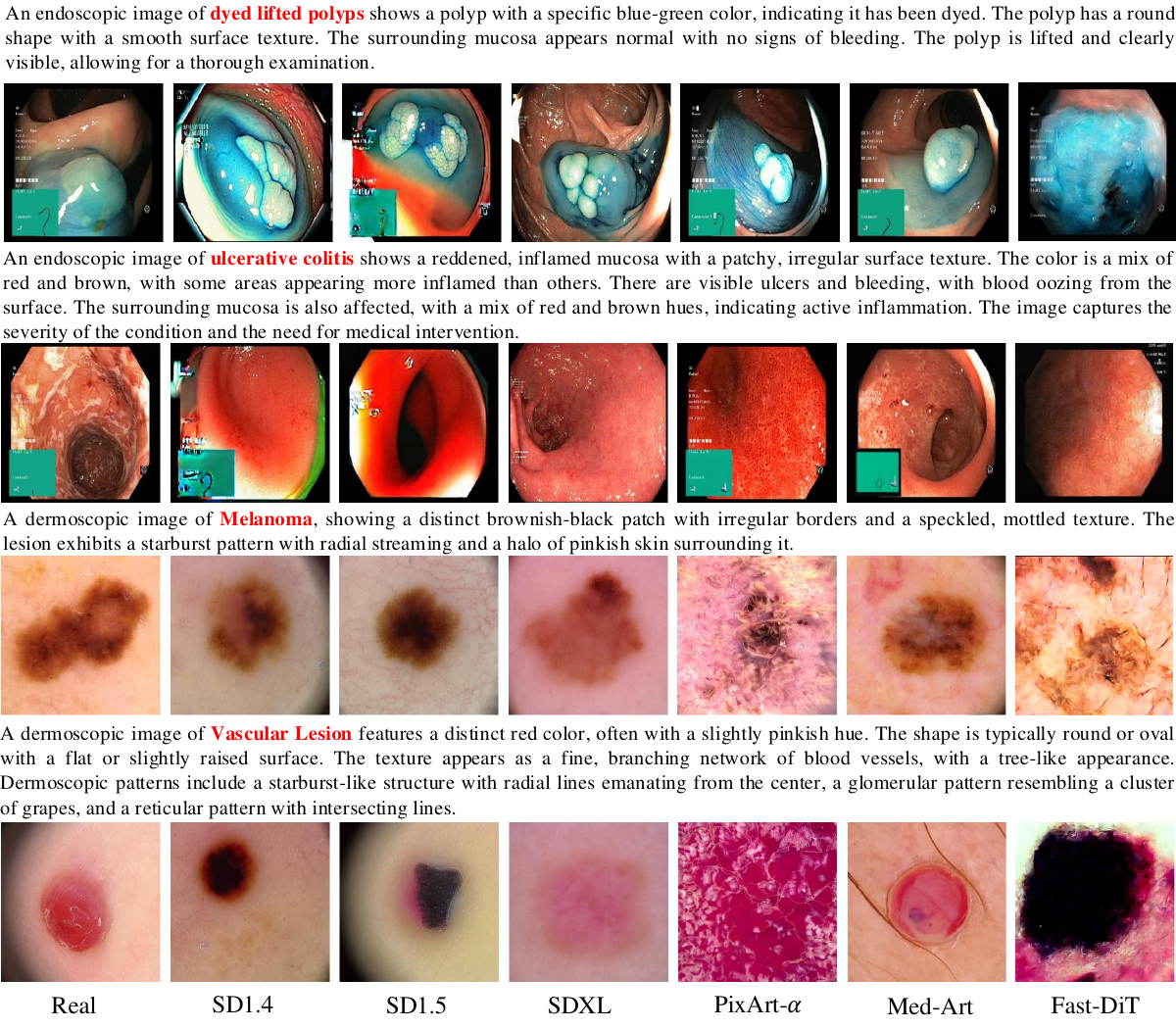}

   \caption{Samples of generated images: each row shows a text prompt and a corresponding image generated by different methods. Fast-DiT generates images using only class labels instead of text prompts. The images generated by the other methods contain unnaturally saturated colors and are less similar to the real image than the result of Med-Art.
   }
   \label{fig:result}
\end{figure}
Even with this relatively small number of steps required, this remains a computationally and memory-intensive process to differentiate. To lower the required memory, at the cost of additional compute, we utilize gradient checkpointing~\cite{chen2016training}. 
%Additionally, the cumulative effect of gradients during these iterations amplifies the impact of pixel-level loss, further increasing the instability of training. 
However, we drastically lower the required computation and training time by adopting an interval optimization strategy, where we only compute the loss every $N$ steps. The loss is then
\begin{equation}
\resizebox{0.8\textwidth}{!}{$
\mathcal{L}_{\text{HLDF}} =
\mathcal{L}_{\text{Pixart-$\alpha$+LoRA}} + 
\begin{cases} 

 \frac{1}{M} \cdot \lcolordist{} , & \text{step} \equiv 0\mod N \\
 0, & \text{otherwise.}
\end{cases}
$}
\end{equation}
Here, \( M \) denotes the number of steps employed by DPM Solver++ for image generation during training, where the default value is 20, and \( \frac{1}{M} \) is the weight.

\begin{table}[h]
    \begin{minipage}{0.52\linewidth}
        \caption{Performance of models \textbf{LoRA}-tuned with image text from VSG on two datasets (except Fast-DiT). Lower is better, best performance indicated in \textbf{bold}.}
        \label{tab:tab2}
        \centering
        \renewcommand{\arraystretch}{1.2} % 增加行高
        \fontsize{8}{10}\selectfont % 设置字体大小为 8pt
        \begin{tabular}{lc c c  c c c} 
            \toprule
            \multirow{2}{*}{Model} & \multicolumn{3}{c}{Kvasir} & \multicolumn{3}{c}{Skin lesions} \\ 
            \cmidrule(lr){2-4} \cmidrule(lr){5-7}
            & FID & KFD  & KID & FID  & HFD & KID \\ 
            \midrule
             Fast-DiT & 102.99 &3.98& 0.059 &86.53& 
             \textbf{3.95} & 0.053 \\
            SD1.4 & 117.85 & 8.36 & 0.077 & 100.75 & 21.21 & 0.091 \\  
            SD1.5 & 143.72 & 6.80  & 0.103 & 88.61 & 14.54 & 0.112 \\  
            SDXL & 125.28 &  7.20 & 0.091 & 103.37 & 26.04 & 0.070 \\ 
            PixArt-${\alpha}$ & 72.86 & 1.85 & 0.030 & 73.68 & 7.14 & 0.045 \\ 
            % Med-Art & \textbf{48.42} & \textbf{1.24} & \textbf{0.015} & \textbf{50.94} & \textbf{5.44} & \textbf{0.027} \\ 
             Med-Art & \textbf{51.99} & \textbf{1.21} & \textbf{0.012} & \textbf{67.45} & 4.47 & \textbf{0.034} \\

            \bottomrule
        \end{tabular}
    \end{minipage}
    \hfill
    \begin{minipage}{0.46\linewidth}
        \caption{The effect of different modules on the Kvasir dataset. Lower is better; best performance indicated in \textbf{bold}.}
        \label{tab:tab3}
        \centering
        \renewcommand{\arraystretch}{1.4}
        \fontsize{8}{8}\selectfont 
        \begin{tabular}{lc c c} 
            \toprule
            Method & FID  & KFD & KID  \\ 
            \midrule
           Empty caption  & 59.12& 12.33 & 0.036\\
           % 126.07 & 47.52 & 0.123 \\
           Only simple caption &55.44  &1.73 & 0.015\\
           % 57.88 & 1.58 & 0.019 \\  
           w/o HLDF (PixArt-${\alpha}$) & 72.86 &1.85  & 0.030 \\ 
            w/o LoRA on T5 & 63.10 & 1.26 & 0.018 \\  
            Hunyuan-DiT \cite{li2024hunyuan}& 143.26 & 15.82 & 0.087 \\  
             with DoRA \cite{liu2024dora} & 55.88 &1.26 & 0.014\\
             % 61.37 & 2.46 & 0.023 \\ 
            Med-Art (Ovis \cite{lu2024ovis})  & \textbf{48.14} & 1.28 & 0.013 \\ 
            Med-Art & 51.99 & \textbf{1.21} & \textbf{0.012} \\ 
            \bottomrule
        \end{tabular}
    \end{minipage}
\end{table}

% This interval strategy significantly reduces the high time and resource costs associated with frequent pixel-level optimization, while alleviating the instability caused by gradient accumulation. By triggering pixel-level loss only at specific intervals, it ensures efficient and stable training, guaranteeing high-quality generation results.
% \begin{table}[h]
%     \caption{Example Table with 4 Images and Corresponding Descriptions}
%     \label{tab:images_with_text}
%     \centering
%     \resizebox{\textwidth}{!}{ 
%         \begin{tabular}{m{0.45\linewidth}<{\centering}|m{0.45\linewidth}<{\centering}} % 每列占宽度的45%
%             \toprule % 顶部粗线
%             \fontsize{8}{8}\selectfont Dyed Lifted Polyps & 
%             \fontsize{8}{8}\selectfont Melanoma \\ 
%             % \cmidrule(lr){1-2} % 分隔线，线条两端带空白
%             \parbox[c][1cm][c]{\linewidth}{\centering \includegraphics[width=1\linewidth]{result1.pdf}} & 
%             \parbox[c][1cm][c]{\linewidth}{\centering \includegraphics[width=1\linewidth]{result2.pdf}} \\ 
%             \midrule % 中间细线
%             \fontsize{8}{8}\selectfont Ulcerative Colitis & 
%             \fontsize{8}{8}\selectfont Vascular Lesion \\ 
%             % \cmidrule(lr){1-2} % 分隔线，线条两端带空白
%             \parbox[c][1.25cm][c]{\linewidth}{\centering \includegraphics[width=1\linewidth]{result3.pdf}} & 
%             \parbox[c][1.25cm][c]{\linewidth}{\centering \includegraphics[width=1\linewidth]{result4.pdf}} \\ 
%             \bottomrule % 底部粗线
%         \end{tabular}
%     }
% \end{table}

\section{Experiments}
% \textbf{Datasets} We evaluate the model on two datasets. The Kvasir dataset contains 8 gastrointestinal endoscopy classes with 1,000 images each~\cite{pogorelov2017kvasir}. The skin lesions dataset consists dermoscopic images
% %, and we use a subset of the data used by Hannemose \etal~\cite{hannemose2022so}
% with multiple diagnostic classes: common nevus (4,906 samples), melanoma (1,180), seborrheic keratosis (805), carcinoma (728), lentigo solaris (229), dermatofibroma (169), vascular lesion (160), and actinic keratosis (140)~\cite{hannemose2022so}. Both datasets are split 80/20 for training and testing per class.
% % Sample images from both datasets are in \cref{table1,fig:result}.

\textbf{Datasets} We evaluate the model on two datasets. The Kvasir dataset contains 8 gastrointestinal endoscopy classes with 1,000 images each~\cite{pogorelov2017kvasir}. The skin lesions dataset includes multiple diagnostic classes, e.g., common nevus (4,906 samples), melanoma (1,180), seborrheic keratosis (805), carcinoma (728), lentigo solaris (229), dermatofibroma (169), vascular lesion (160), and actinic keratosis (140)~\cite{hannemose2022so}. Both datasets are split 80/20 for training and testing per class.

\noindent \textbf{Baselines} We compare against state-of-the-art text-to-image diffusion models, including U-Net-based~\cite{ronneberger2015u} models from the Stable Diffusion~\cite{rombach2022high} family (SD1.4, SD1.5, and SDXL~ \cite{podell2023sdxl}), as well as Hunyuan-DiT~\cite{li2024hunyuan} and PixArt-${\alpha}$. 
We LoRA fine-tune both the denoising network and the text encoder for all models, except for Hunyuan-DiT, which only supports LoRA fine-tuning for the denoising network.
Finally, we compare to Fast-DiT~\cite{jin2024fast}, a class-conditional diffusion model, using the DiT-S/8 variant trained from scratch, conditioned on class labels.

% \begin{table}[h]
%     \caption{Comparing with Other SOTA T2I Methods, Kvasir}
%     \label{tab1}
%     \centering % Center the table
%     { % 将表格调整为适应文本宽度
%         \begin{tabular}{c | c c c c c c } % 设置列数与格式
%             \toprule % 加粗的顶部横线
%             \fontsize{10}{12}\selectfont % 设置字体大小为10pt，行距为12pt
%             Models &  FID $\downarrow$ & Kvasir-FD $\downarrow$ & KID $\downarrow$  \\ 
%             \midrule % 中间横线

%             Med-Art  (M=20,steps=2000,w2=0.1) & 55.25 &  1.35 & 0.016 \\
%              Med-Art  (M=20,steps=1000,w2=0.1) & 56.04 &  1.48 & 0.018 \\
%                       Med-Art  (M=20,steps=750,w2=0.1) & 54.41 &  2.07 & 0.015 \\
%                       Med-Art (M=20,steps=500,w2=0.1) & 48.42 &  1.24 & 0.015 \\
%            Med-Art  (M=20,steps=250,w2=0.1) & 44.23  & 1.66 &  0.009\\
%             \midrule
%             Med-Art (M=20,steps=500,w2=0.1) & 48.42 &  1.24 & 0.015 \\
%              Med-Art  (M=20,steps=500,w2=0.05) & 51.99 &  1.21 & 0.012 \\
%             Med-Art  (M=20,steps=500,w2=0.15) & 
%             49.77 &1.27& 0.011
%             % Med-Art (M=10,steps = 500, w2=0.1) & \textbf{48.04} &  \textbf{2.03} & \textbf{0.014} 
%             \\  
%             \bottomrule % 加粗的底部横线
%         \end{tabular}
%     } % 结束 resizebox
% \end{table}

\noindent \textbf{Experimental Setup} We train all models on a single A100 GPU using mixed-precision \cite{micikevicius2017mixed}. All models are trained to generate \(512 \times 512\) images with a learning rate of \( 10^{-4}\). The LoRA rank \(8\) and batch size \(1\). We train for 10 epochs on skin lesions and 15 epochs on Kvasir datastet. The pixel-level loss \cref{eq:colordist} is applied every \(N = 500\) steps for the Kvasir dataset and every \(N = 1500\) steps for the skin lesions dataset. 
% A random subset of training texts, matching the test set size (\(1600\) for Kvasir and \(1665\) for skin lesions), serves as text prompts for generation.
A random subset of training images, matching the test set size (1600 for Kvasir and 1665 for skin lesions), is used to generate text prompts for test image synthesis.

\noindent \textbf{Generative performance}
In addition to the Fréchet Inception Distance (FID) \cite{heusel2017gans}, we replace the Inception V3-based feature extractor with publicly available ViT models\footnote{
\href{https://huggingface.co/mmuratarat/kvasir-v2-classifier}{mmuratarat/kvasir-v2-classifier}, \href{https://huggingface.co/Anwarkh1/Skin_Cancer-Image_Classification}{Anwarkh1/Skin\_Cancer-Image\_Classification}}, pretrained on the Kvasir and the HAM10000 datasets, terming these variants “KFD” and “HFD” respectively. Additionally, we use Kernel Inception Distance (KID)~\cite{binkowski2018demystifying}, which enhances sensitivity for small datasets. \Cref{tab:tab2,fig:result} present the quantitative analysis and generated medical image examples from different models. For all models, Med-Art achieves the best performance across all metrics, except HFD. While PixArt-$\alpha$ ranks second, it suffers from severe color distortion, exhibiting oversaturation on the Kvasir dataset and even color deviation in skin lesion images. SD1.4 and SD1.5 typically produce overly smooth or distorted images, especially on the Kvasir dataset, while SDXL, although improving colors on Kvasir, still generates blurry and distorted skin images. Qualitatively, Fast-DiT struggles to generate features close to real pathological images. Med-Art generates highly realistic and high-quality medical images with superior color fidelity that closely resemble real medical imagery.

\begin{table}[hbt]
\newcommand{\second}[1]{\color{gray}\textbf{#1}}
    \caption{Classification performance of models trained on synthetic or real data. Higher is better, best performance indicated in \textbf{bold}, second best in \second{gray}.}
    \label{tab4}
    \centering
    {
    
        \begin{tabular}{c  c  c c c  c c c }
            \toprule
            \fontsize{8}{8}\selectfont 
            \multirow{2}{*}{\centering Training data} & \multirow{2}{*}{\centering Method} & \multicolumn{3}{c}{Kvasir} & \multicolumn{3}{c}{Skin lesions} \\ 
            \cmidrule(lr){3-5}\cmidrule(lr){6-8}
            & & F1 & BACC & AUC & F1 & BACC & AUC \\ 
            \midrule
            \multirow{6}{*}{Synthetic} & 
            Fast-DiT & 0.6492 &0.6512 & 0.9256 &  0.5696 & 0.2287 & 0.7920 \\
            &SD1.4 & 0.5957 &0.6450 & 0.9368 &0.5177  & \textbf{0.3464} & 0.7840 \\  
            & SD1.5 &  0.6192&0.6669 & 0.9622 & 0.5946 & 0.3039 & \textbf{0.8132} \\  
            & SDXL & \second{0.6830}&\second{0.7131}&0.9683  &0.3866  & 0.2721 &  0.7348\\  
            & PixArt-${\alpha}$ &  0.6291&0.6375 & \second{0.9730} &   0.5154& 0.1680 &  0.7541 \\  
            % & Med-Art &   \textbf{0.7932}& \textbf{0.7906}  &\textbf{0.9759}& \textbf{0.5813}  & 0.5742& \textbf{0.8143} 
            & Med-Art &   \textbf{0.8072}& \textbf{0.8119}  &\textbf{0.9847}& \textbf{0.6090}  & \second{0.3182} & \second{0.7994}
            \\
            \midrule
            Real & -  &0.9361 & 0.9363  &0.9962  &  0.7494 & 0.5080 & 0.9222 \\ 
            
            \bottomrule
        \end{tabular}
    }
\end{table}

\Cref{tab:tab3} illustrates the impact of different modules on Med-Art's performance on the Kvasir dataset. Compared to the complete Med-Art model, removing VSG (using empty caption or simple caption), HLDF (equivalent to PixArt-$\alpha$), or LoRA on T5 all result in reduced performance. These findings indicate that the detailed, context-aware text prompts generated by VSG play a critical role in boosting model performance. Meanwhile, HLDF introduces only moderate computational overhead, increasing training time by 9\% on the Kvasir dataset ($N=500$, 12h14m~$\rightarrow$~13h20m) and 6\% on the skin lesions dataset ($N=1500$, 9h9m~$\rightarrow$~9h43m) in our experiments. More importantly, as demonstrated by the visual comparison between PixArt-$\alpha$ and Med-Art in \cref{fig:result}, it significantly improves image quality—especially by reducing color distortion—without adding any extra time during the generation process. Considering the substantial gains in generation fidelity and clinical applicability (\cref{tab:tab3}, \cref{fig:result}), this trade-off is reasonable and worthwhile.
% Meanwhile, HLDF is also essential for model optimization, and this is further evidenced by the visual comparison between PixArt-$\alpha$ and Med-Art in \cref{fig:result}, which demonstrates its effectiveness in mitigating color distortion.
Furthermore, fine-tuning the T5 encoder significantly enhances image generation; even without it, Med-Art outperforms the state-of-the-art Hunyuan-DiT, which fine-tunes only DiT. Although the recently introduced DoRA~\cite{liu2024dora} demonstrates competitive performance, our experiments show that LoRA consistently outperforms DoRA on the Med-Art, reaffirming LoRA as the optimal choice for our setting. To demonstrate that our model generalizes to other prompts, we show results using prompts generated by Ovis~\cite{lu2024ovis}, which achieved very similar results.

% Furthermore, fine-tuning the text encoder substantially enhances image generation quality. Even without text encoder fine-tuning, Med-Art still outperforms the current state-of-the-art Hunyuan-DiT model, which is only fine-tuned on DiT. Although the recently introduced DoRA~\cite{liu2024dora} demonstrates competitive performance, our experiments show that LoRA consistently outperforms DoRA on the Med-Art task, reaffirming LoRA as the optimal choice for Med-Art.

\noindent \textbf{Classification performance}
Finally, to eliminate the influence of real data on classification performance, instead of using generated images for data augmentation \cite{yuan2024adapting}, we fine-tune a classification model directly with them and evaluate its performance on the real test set. All images in the training set are used to generate texts and training data. We use a Vision Transformer~\cite{alexey2020image}, ViT-B/16, that is pretrained on natural images. We fine-tune using the AdamW optimizer~\cite{loshchilov2017decoupled} for five epochs with a learning rate of $2\cdot 10^{-5}$. \Cref{tab4} shows that images generated by Med-Art enable the best classification performance on the Kvasir dataset. For the skin lesions dataset, while SD1.4 achieves the highest balanced accuracy (BACC) and SD1.5 obtains the best AUC, both models underperform Med-Art on the other two metrics. Taking into account all metrics, Med-Art demonstrates the best overall classification performance, not too far from the results obtained using real data. This highlights Med-Art’s strong generative capability in producing category-specific samples that most resemble real images.

% \noindent \textbf{Trade-off Between Efficiency and Performance of HLDF} HLDF incurs additional computation by invoking DPM-Solver++ every \(N\) steps to generate supervision images. Under single-GPU mixed-precision training, this overhead is modest: training time increases by about 9\% on Kvasir (\(N=500\), 12h14m to 13h20m) and 6\% on Skin Lesions (\(N=1500\), 9h9m to 9h43m). As Table~3 and Figure~2 show, HLDF substantially enhances image quality and color consistency, a key requirement for clinical use. Thus, this moderate overhead is justified to ensure diagnostic validity of generated images.
\section{Conclusion}
This study introduces Med-Art, which achieves state-of-the-art text-to-image generation in medical settings with limited numbers of images and text data by fine-tuning large-scale generative models on small medical image datasets, while also effectively mitigating color distortion in the generated images. Med-Art's use vision-language models to overcome the lack of medical texts, furthermore implies that it can improve further upon future advancements in these.
%Med-Art is notably dependent on VLMs, but with the advancement of large multimodal models, we believe Med-Art will pave the way for the integration of artificial general intelligence in the medical domain.
\bibliographystyle{splncs04}
\bibliography{references}

\end{document}